\RequirePackage[2020-02-02]{latexrelease}
\documentclass{article}
\usepackage{spconf,amsmath,graphicx}
\usepackage[dvipsnames]{xcolor} 
\usepackage{url}
\usepackage{amsfonts} 
\usepackage{hyperref}

\usepackage[printwatermark=true]{xwatermark}
\newwatermark[page=1,color=gray!60,fontfamily=cmr,fontseries=m,scale=0.3,xpos=0,ypos=-125, width=2.7\paperwidth,align=left]{\textcopyright\ 2022 IEEE. Personal use of this material is permitted. Permission from IEEE must be obtained for all other uses, in any current or future media, including reprinting/republishing this material for advertising or promotional purposes, creating new collective works, for resale or redistribution to servers or lists, or reuse of any copyrighted component of this work in other works.}

\def\LN{{\text{LN}}}
\def\MSA{{\text{MSA}}}
\def\SA{{\text{SA}}}
\def\MLP{{\text{MLP}}}

\title{Exploiting spatial sparsity for event cameras with visual transformers}
\name{Zuowen Wang$^\dag$\qquad Yuhuang Hu$^\dag$\thanks{\dag Equal contribution.}\qquad Shih-Chii Liu\thanks{This work was partially funded by the European Union's Horizon 2020 research, innovation programme under grand agreement No 899287, and the Swiss National Competence Center in Robotics (NCCR Robotics).}}

\address{Institute of Neuroinformatics, University of Z\"urich and ETH Z\"urich,
Z\"urich, Switzerland}

\begin{document}
%
\maketitle
\begin{abstract}
Event cameras report local changes of brightness through an asynchronous stream of output events. Events are spatially sparse at pixel locations with little brightness variation. We propose 
using a visual transformer (ViT) architecture 
to leverage its ability to process a variable-length input. 
The input to the ViT consists of events that are accumulated into time bins and spatially separated into non-overlapping sub-regions called patches. 
Patches are selected when the number of nonzero pixel locations within a sub-region is above a threshold. We show that by  fine-tuning a ViT model on these selected active patches, we can reduce the average number of patches fed into the backbone during the inference 
by at least 50\% with only a minor drop (0.34\%) of the classification accuracy on the N-Caltech101 dataset. This reduction translates into a decrease of 51\% in Multiply-Accumulate (MAC) operations and an increase of 46\% in the inference speed using a server CPU.
\end{abstract}


%
\begin{keywords}
Event cameras, Spatial sparsity, Reduced computation, Visual transformers
\end{keywords}

\section{Introduction}
\label{sec:intro}

The use of event cameras such as the Dynamic Vision Sensor (DVS) \cite{dvs} and the Dynamic and Active Pixel Vision Sensor (DAVIS) \cite{LiColorDavis2015} in computer vision research has been growing in recent years~\cite{GallegoEventVision2022}. Event cameras generate asynchronous events from pixels that see a local change in brightness. 
An event is only generated once the log intensity change surpasses a threshold, thus 
the output is spatially sparse in visual scenes where there are few moving objects.

Deep neural networks (DNNs) have been used together with event cameras to solve vision tasks such as 
object recognition and optical flow estimation \cite{GallegoEventVision2022, evflownet, e_raft}. Convolutional neural networks (CNNs) are frequently the choice of network architecture.
However, these methods often require an event preprocessing step, which accumulates a batch of events and transforms them into an event tensor that could be used as the input of the CNN.
Regular CNNs have to process the whole input frame unless special hardware is used~\cite{nullhop}. The CNNs also lose the sparsity in higher level feature maps, thus diminishing the advantage of spatial sparsity of events. 

\begin{figure}[t]
\centering
\includegraphics[width=0.95\linewidth]{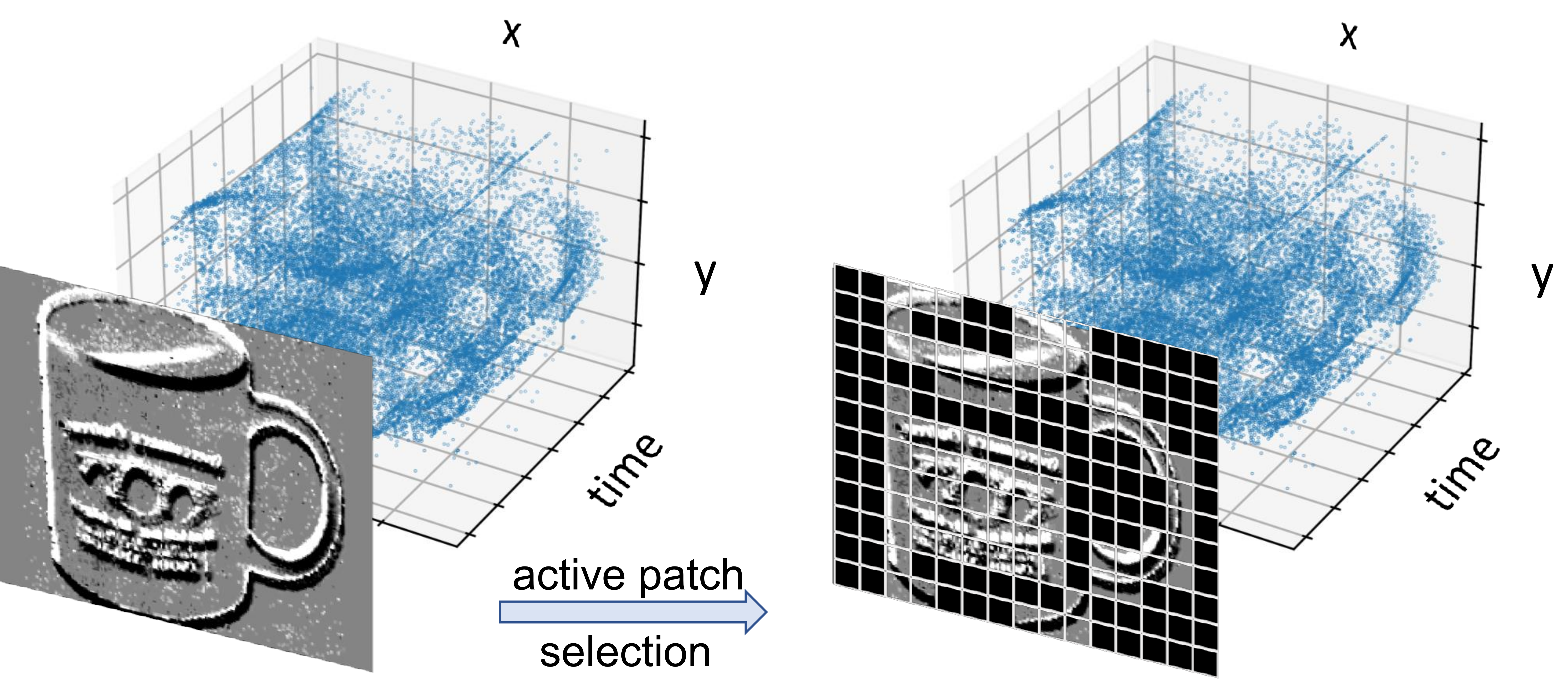}
\caption{An illustration of the active patch selection process. Only patches with enough nonzero pixel values will be included during the inference.}
\label{fig:active_patch}
\end{figure}

\begin{figure*}[t]
\includegraphics[width=1.0\textwidth]{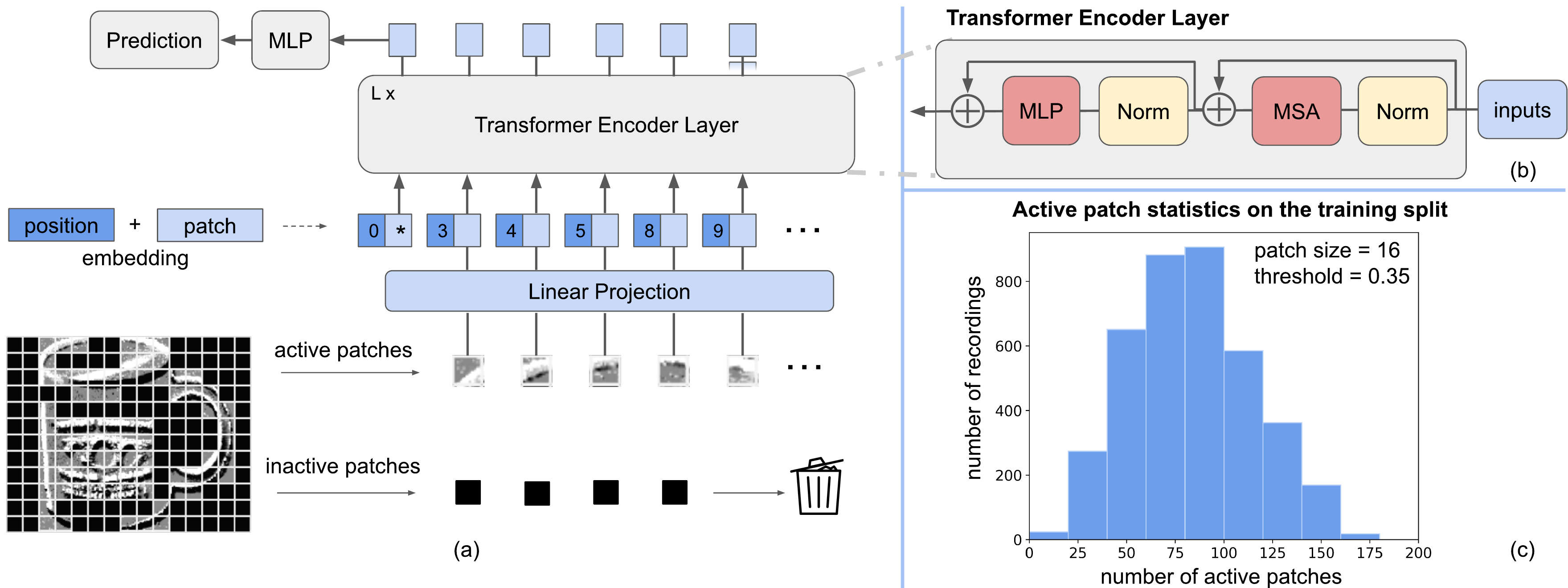}
\caption{(a) An illustration of the ViT architecture and the active patch preprocessing pipeline. During inference, an input event frame is divided into active and inactive patches. 
The active patches will first be flattened and linearly projected then added with positional embeddings. 
(b) Building blocks of the transformer encoder layer. The two major computational bottelnecks are marked with red. (c) The histogram of number of active patches for the event frames in the training split of N-Caltech101 dataset. By setting the active patch threshold to 0.35, the number of active patches processed by the transformer is reduced by more than 50\% on average. 
}
\label{fig:vit_dvs}
\end{figure*}

Recently, a new DNN architecture, \emph{i.e.}, visual transformers (ViT) \cite{dosovitskiy2020vit}, has been gaining a lot of attention in vision research. Its variants have achieved state-of-the-art accuracy across various tasks~\cite{transformer_survey}. A ViT model first divides an input image into $n$ sub-regions (patches), then embeds each patch into a $D$-dimensional vector. Despite its great success, the self-attention (SA) module, which is repeatedly used as a building block in the transformer architecture, has quadratic memory and time complexity with respect to the number of input patches. This dependence hinders the deployment of transformers on memory and compute-constrained hardware devices. Works have been proposed to approximate the original attention matrix with lower complexity functions~\cite{Reformer,linformer, NystrmformerAN}. However, the fully connected layers have a larger time complexity than the SA module when input sequence lengths are values within the range for practical applications in vision tasks. 

Little work has been proposed on using ViTs with event camera inputs. The event data naturally provides  `regions of interest' through event responses, particularly in scenes with high spatial frequency textures or scenes with fast moving objects. In this work, we capitalize on the spatial sparsity offered by the event camera in combination with the visual transformer.
We use an `active patch' selection procedure to filter out ``inactive" areas and therefore reduce the number of input patches.
We show that ViTs can maintain the original accuracy on a classification task even with on average half the number of input patches.
We further determine the computation savings from this procedure and study the trade-off between the speed-up of the inference and the classification accuracy. We compare our work with other frame-based CNN models.

\section{Methodology}
\label{sec:methodology}
In this section, we first introduce the ViT architecture used in our study and especially its flexibility of handling variable input length. In Sec. \ref{subsec:preprocessing_active_patch} we describe the details of the event preprocessing pipeline and the active patch selection, which is vital for exploiting the spatial sparsity of event data and the accompanying computation reduction. In Sec. \ref{subsec:time_complexity} we analyse the time complexity of the ViT and its dependence on the input sequence length. We validate this analysis with the experimental results in Sec. \ref{sec:experiments}.

\subsection{Visual transformer}\label{subsec:dvs_vit}
For the ViT backbone \cite{dosovitskiy2020vit}, the selected frame patches are first projected with an embedding $\textbf{E} \in \mathbb{R}^{(P^2\cdot C)\times D}$, where $P^2$ is the patch size, $C$ is the number of patch channels and $D$ is the embedding dimension for each patch. As shown in Eq.~\ref{eq:embed}, in the first layer of the DVS ViT, the $n$ flattened patches $x_{1,...,n}$ are embedded as:
\begin{align}\label{eq:embed}
    z_{0} = [x_{\text{class}}; x_{1}\textbf{E};x_{2}\textbf{E}; \cdots ;x_{n}\textbf{E}] + \textbf{E}_{\textit{pos}},
\end{align}
where $\textbf{E}_{\textit{pos}} \in \mathbb{R}^{(n+1)\times D}$ is the positional embedding matrix and $x_{\text{class}} \in \mathbb{R}^{D}$ is the additional class embedding. Then the embedded vector $z_0 \in \mathbb{R}^{(n+1)\times D}$ is passed into the $L$ transformer encoder layers.

The two most important components in each transformer encoder layer are multi-head self attention (MSA) and multi-layer perception (MLP), which are also computationally heavy. The $l$-th transformer encoder layer shown in Fig.~\ref{fig:vit_dvs} (b) can be written as:
\begin{align}
    z'_{l} &= \MSA(\LN(z_{l})) + z_{l} \\
    z_{l+1} &= \MLP(\LN(z'_{l})) + z_{l}^{'} \label{eq:mlp} ,
\end{align} where $l\in \{1, 2, ..., L\}$ and $\LN(\cdot)$ is the layer normalization function. 

The MSA consists of $k$ individual self-attention (SA) heads. Each SA head is formulated as:
\begin{align} \label{eq:SA}
    \SA(z_l) &= \text{softmax}(\textbf{qk}^{T}/\sqrt{D_h})\cdot \textbf{v}, \quad \text{where}\\
    [\textbf{q}, \textbf{k}, \textbf{v}] &= [z_{l}\textbf{U}_q, z_{l}\textbf{U}_k, z_{l}\textbf{U}_v]. \label{eq:qkv}
\end{align}
The three learnable matrices $\textbf{U}_q,\textbf{U}_k,\textbf{U}_v \in \mathbb{R}^{D\times D_h}$ projects embedding $z_l \in \mathbb{R}^{(n+1)\times D}$ to $\mathbb{R}^{(n+1)\times D_h}$, where $D_h$ is a dimension of our choice. Thus, each SA head is in $\mathbb{R}^{(n+1)\times D_h}$. The $k$ SA heads operate individually on the embedded patches and then their outputs are concatenated and projected back to $\mathbb{R}^{D}$ by the trainable projection $\textbf{U}_{\MSA}\in\mathbb{R}^{k\cdot D_h \times D}$ in the MSA, formulated as:
\begin{align} \label{eq:MSA}
\MSA(z_l) = [\SA_1(z_l); \SA_2(z_l); ...; \SA_k(z_l)]\textbf{U}_{\MSA}.
\end{align} 

Each MLP module consists of two layers of fully connected layers with GELU non-linearity~\cite{gelu} in our experiments. We denote the hidden dimension of the MLP as $D_{\textit{mlp}}$. The output layer should have dimension $D$ since it has to align with the next encoder layer. The fully connected layers operate patch-wise, namely, they are applied $(n+1)$ times in each transformer encoder layer.

Notice that the dimensionalities of inputs and outputs of MSA module and MLP layers are both $(n+1)\times D$ and $n$ is a variable parameter depending on the number of input patches.

\subsection{Event preprocessing and active patch selection} \label{subsec:preprocessing_active_patch}
We conduct our study on a classification task using the N-Caltech101~\cite{n_caltech} dataset. It contains in total 8,246 event recordings, each with one label out of 100 classes. The entire dataset is split into 45\% for the training set, 30\% for the validation set and 25\% for the testing set. 

We construct event frames using a event voxel grid representation~\cite{zhu2019unsupervised}. The voxel grid is built with bilinear interpolation along the time axis and accumulated into $C$ channels. More specifically, for a given sequence of events $\mathcal{E}$, it is formulated as:
\begin{equation}
   \mathcal{E}= \{e_i\}_{i=1}^{N} = \{x_i,y_i,t_i,p_i\}_{i=1}^{N}. 
\end{equation} 
Each event $e_i$ is depicted by its spatial coordinates $(x_i, y_i)$, time stamp $t_i$ and polarity $p_i \in \{\pm 1\}$ and the recording $\mathbb{E}$ contains $N$ events in total. 
The entire event recording is evenly divided into $C-1$ temporal bins with bin size $\Delta T = \frac{t_N - t_1}{C-1}$. Each channel will accumulate the events in the two temporally nearby bins with time distance weight $(1-\frac{|t_i - t'_c|}{\Delta T})$, where $t'_c$ is the time stamp of the $c$-th channel to accumulate events into. The value $I(x, y, c)$ at spatial location $(x, y)$ and channel $c$ in the event frame is:
\begin{equation}\label{eq:1}
\begin{split}
    &I(x, y, c) \\ &=\sum_{t'_{c-1}<t_i\leq t'_{c+1}} (1-\frac{|t_i - t'_c|}{\Delta T})\delta(x - x_i, y - y_i)p_i , 
\end{split}
\end{equation}
where $t'_c = t_1 + (c-1) \Delta T$ for $c\in\{1,...,C\}$ and $t'_0 = t_1$, $t'_{C+1} = t_N$ give the boundary conditions. The $\delta(a,b)$ function:
\begin{equation}
        \delta(a, b) =     \begin{cases}
      1 & \text{if } a=0 \text{ and } b=0\\
      0 & \text{otherwise}
    \end{cases} 
\end{equation}
is only evaluated to one only when an event in the given time bins matches the spatial coordinate $(x, y)$. 

We then resize and pad the event frame with zeros to $192 \times 240$ so that the resulted event frames keeps the original aspect ratio and pads to the same size. During training, the data was augmented with random horizontal flips and affine transformations with a maximum 15$^o$ of rotation and 10\% of translation. Nonzero entries in the event frame are normalized with the mean and variance calculated based on these entries.

We divide the event frame spatially into non-overlapping sub-regions. 
We use a patch size of $16 \times 16$ which divides the $192 \times 240 \times C$ into 180 sub-regions of size $16 \times 16 \times C$. In our experiments, we use $C=9$ since it shows the best classification accuracy in the grid search. 

We distinguish the active patches and inactive patches with the following simple statistics of events. We calculate the number of nonzero entries in each event frame patch. Then the active ratio (AR) is defined as $\text{AR} = \frac{\text{number of non-zero entries}}{\text{total entries}}$. The total entries in each patch is $(16 \times 16 \times 9)$. For each patch, if its AR is larger than a predefined threshold, then we regard it as an active patch and they will be fed into the transformer backbone for making a prediction. Otherwise, it is an inactive patch and will be discarded during inference or training. As shown in Fig.~\ref{fig:vit_dvs} (c), for the training split of the N-Caltech101 dataset, with a threshold 0.35, the average number of patches fed into the transformer will be reduced to less than 50\% with the active patch selection.
 
After patch selection, each active patch will be projected into a $D$-dimensional vector as in Eq.~\ref{eq:embed}. We still maintain a full-sized positional embedding weight matrix. The positional embeddings added to the patch embeddings are selected according to the positions of the active patches. This is to keep the information about positions among the patches. 


\subsection{Computational bottleneck analysis} \label{subsec:time_complexity}
Here we analyse the number of floating point operations (FLOPs) of the two computational bottlenecks of transformers, namely, the MSA module and the MLP layer, to show the benefits of reducing the number of patches in the input sequence.
Following notations in Eq.~\ref{eq:qkv}, assume the input length is $n$ (omitting the additional class embedding in the following calculation for simplicity), we need approx. $6n\cdot D \cdot D_{h}$ FLOPs to compute the $\textbf{q}, \textbf{k}, \textbf{v}$ matrices in each attention head. In Eq.~\ref{eq:SA}, approx. $2D_{h} \cdot n^{2} + n^{2}$ operations are needed for $\textbf{q}\textbf{k}^{T}$ and scaling with $\sqrt{D_h}$ respectively. The softmax function and matrix multiplication with $\textbf{v}$ requires $2\cdot n^{2} + 2D_{h} \cdot n^{2} $ operations. Thus, in total $k$ self-attention heads and the $\mathbf{U}_{\MSA}$ projection consumes approx. $k(3+4D_{h})n^{2} + 8kDD_{h}\cdot n$ FLOPs. The FLOPs needed for the patch-wise MLP in Eq.\ref{eq:mlp}
costs $4D D_{\MLP} \cdot n$ FLOPs.

Considering the ViT backbone used in this work,
we have $k=12, D_{h} = 64, D = 768, D_{\MLP} = 3072$. We need $(3108\cdot n^2 + 4718592\cdot n)$ FLOPs for MSA and $(9437184 \cdot n)$ FLOPs for MLP. The FLOPs of MLP module dominates when the number of input patches is smaller than 1500, which includes our use cases. This indicates that cutting down the number of inputs will induce a linear FLOPs reduction and is overall beneficial to the computational speed. Stand-alone efforts of replacing the SA with sub-$\mathcal{O}(n^2)$ approximations~\cite{Reformer,linformer, NystrmformerAN} is less effective for ViTs.

\section{Experiments}
\label{sec:experiments}
In this section, we first introduce our detailed experimental settings in Sec.~\ref{sec:exp_set} for 
DVS-ViT training and inference. In Sec.~\ref{sec:exp_results} we show the model performance and computation reduction with different active patch thresholds. 
We evaluate the models on the N-Caltech101 dataset~\cite{n_caltech}. The event frames per second values are measured solely on an Intel Xeon W-2195 CPU.

\subsection{Experimental settings}
\label{sec:exp_set}
In this section, we describe the details of the experimental settings. The choices of architecture-related hyperparameters in the experiments are described in Sec.\ref{subsec:time_complexity}.

For the experiments in Fig.~\ref{fig:acc_mac_threshold}, we first fine-tune an ImageNet pretrained ViT backbone on the N-Caltech101 dataset to get a well-performing baseline model (active patch threshold $=0.0$). We name this baseline model DVS-ViT.

We then further fine-tune the DVS-ViT baseline model with different active patch thresholds ranging from 0.05 to 0.7 with step size 0.05 separately. In both training and evaluation of the models, we use the same active patch threshold in default cases. We also test a `mixed' training procedure which samples a threshold uniformly at random between 0.0 to 0.7 during training for each iteration, and evaluated with a threshold 0.35. We find that it achieves a good trade-off between accuracy and MACs reduction (marked with the golden star in Fig.~\ref{fig:acc_mac_threshold}). We name the models trained with active patches as DVS-ViT-$\{\text{threshold},\text{mixed}\}$. Due to the constraint of gradient computation of different input lengths, the training batch size is set to 1. 

All models are trained for 50 epochs and are evaluated on the checkpoints with the highest validation accuracy. Each model is run three times with different random seeds to compute the mean and standard deviation. All training experiments use the AdamW optimizer~\cite{adamW} and the cross-entropy loss. The training of the baseline model uses batch size 16 and an initial learning rate of 5e-5. The training of the active patches uses an initial learning rate of 1e-5. 

\subsection{Experimental results}
\label{sec:exp_results}
\begin{figure}[t] 
\centering
\includegraphics[width=1.0\linewidth]{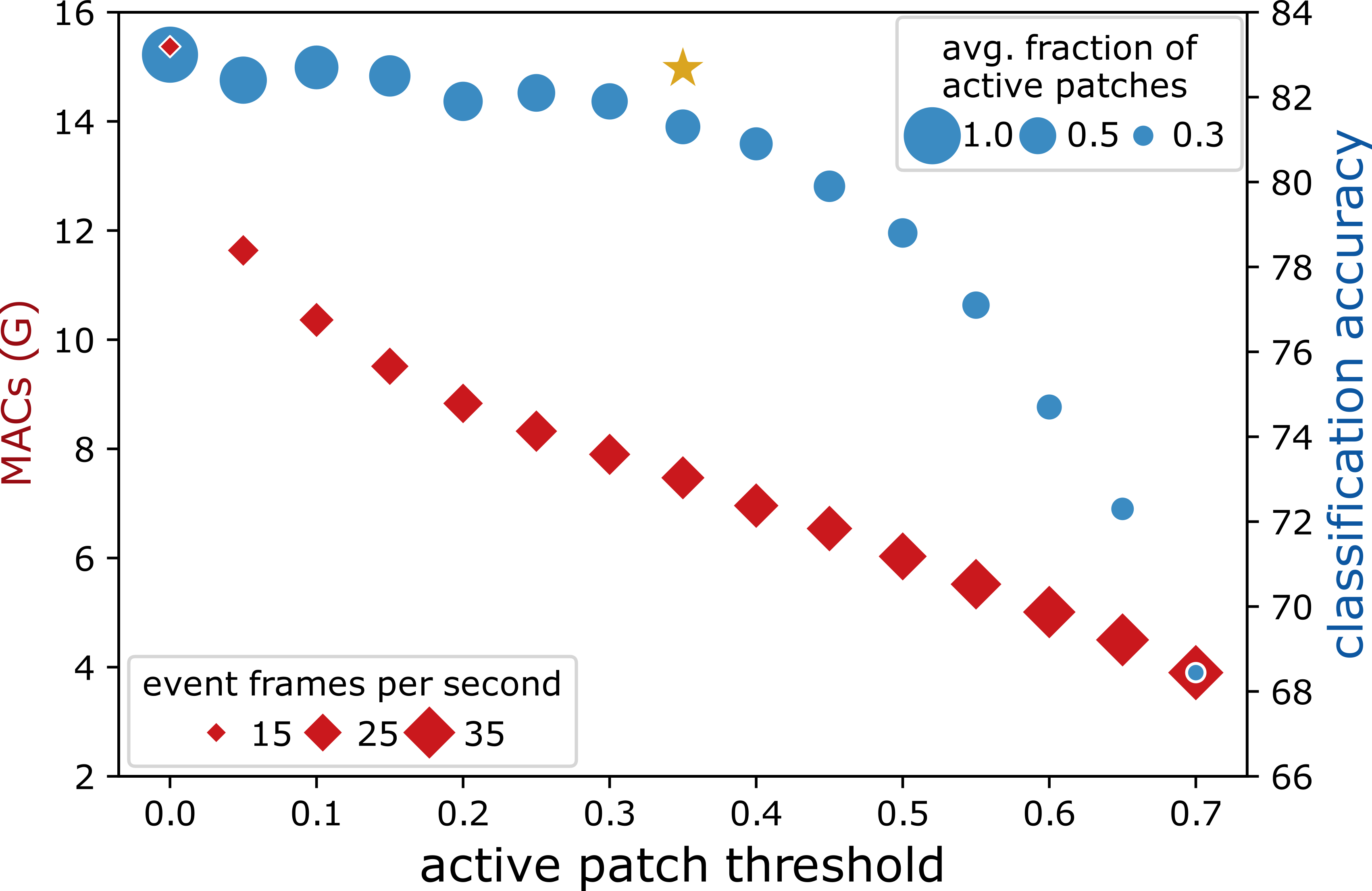}
\caption{Active patch training and inference with different thresholds. Red diamond points show the average inference MACs and blue circles show the accuracy at different threshold levels. The golden star model is trained using active patch thresholds randomly selected from $[0.0, 0.7]$.  Reported inference accuracy is for a 0.35 threshold.}
\label{fig:acc_mac_threshold}
\end{figure}

Fig.~\ref{fig:acc_mac_threshold} shows the classification accuracy and MACs needed for models with different active patch thresholds. By setting the threshold to 0.35, we achieved a 51.4\% MACs reduction with a minor accuracy drop. Moreover, by adopting the `mixed' training procedure, we achieved the same accuracy as the baseline model while cutting the computation in half. 

The sizes of the blue circles show the average fraction of active patches which equals to the average number of active patches in comparison with the total number of patches. With roughly 0.5 average fraction (at threshold 0.35), we achieved 51.4\% MACs cut, which verifies the analysis in Sec.~\ref{subsec:time_complexity}.

We also compare our results with other works \cite{hats, eventDrop, est} in Tab.\ref{tab:comp_others}. Methods including learnable event representation formulation with well-studied CNN backbones. We showed that pretrained ViTs could achieve similar or higher performance as pretrained CNNs on the N-Caltech101 dataset, despite the known issue of requiring more data than CNNs to generalize well ~\cite{vit_small_dataset} due to lack of convolutional inductive bias which could benefit vision related learning tasks.

\begin{table}[h]
\centering
    \caption{Performance comparison with other works.}
    \label{tab:comp_others}
\begin{tabular}{lccc}
 \hline
  \multicolumn{1}{c}{Model} & Accuracy (\%)  & Params & MACs \\
 \hline
 
 HATS \cite{hats} & 64.2  &  -&- \\ 
 VGG-19 \cite{eventDrop} & 76.6 $\pm$ 0.8  & 139.98M & 18.05G \\
 ResNet34 \cite{eventDrop}  & 82.5 $\pm$ 0.8  & 21.33M & 3.41G\\
 EST \cite{est} & 81.7   & 21.38M & 5.12G \\
 \hline
 DVS-ViT &  83.00 $\pm$ 0.36  & 85.69M &15.37G \\
 DVS-ViT-0.1 &  82.72 $\pm$ 0.21  & 85.69M &10.36G \\
 DVS-ViT-0.35 & 81.27 $\pm$ 0.53 & 85.69M &7.47G  \\
DVS-ViT-mixed & 82.66  $\pm$ 0.34 & 85.69M & 7.47G \\

\hline
\end{tabular}
\end{table}

\section{Conclusion}
\label{sec:conclusion}
In this work, we proposed a pipeline using the ViT as the backbone for processing event camera data. The pipeline exploits the spatial sparsity of events with the help of the ability of ViTs to process variable length inputs. With the active patch selection process, we reduced the number of FLOPs of ViT by more than 50\% on the N-Caltech101 dataset while still maintaining a competitive accuracy compared to previous state-of-the-art CNN based methods.

\vfill\pagebreak

\bibliographystyle{IEEEbib}
\bibliography{refs}

\clearpage
\end{document}